\documentclass{article}
\usepackage{arxiv}
\usepackage[utf8]{inputenc} % allow utf-8 input
\usepackage[T1]{fontenc}    % use 8-bit T1 fonts
\usepackage{hyperref}       % hyperlinks
\usepackage{url}            % simple URL typesetting
\usepackage{booktabs}       % professional-quality tables
\usepackage{amsfonts}       % blackboard math symbols
\usepackage{nicefrac}       % compact symbols for 1/2, etc.
\usepackage{microtype}      % microtypography
\usepackage{times}
\usepackage{latexsym}

\usepackage{amsmath}
\usepackage{amsfonts}
\usepackage{amssymb}
\usepackage{wrapfig}
\usepackage{subcaption}
\usepackage{multirow}
 \usepackage{mathtools} 
\usepackage{verbatim}
\usepackage{anyfontsize}
\usepackage{color}
\newcommand{\distas}[1]{\mathbin{\overset{#1}{\kern\z@\sim}}}%
\usepackage{enumitem}
\usepackage[lined,boxed,commentsnumbered,ruled,linesnumbered]{algorithm2e}
\usepackage{booktabs} % toprule,midrule
\usepackage{colortbl} % rowcolor
\usepackage{authblk}
\definecolor{Gray}{gray}{0.93}

\newcommand{\RN}[1]{%
	\textup{\lowercase\expandafter{\it \romannumeral#1}}%
}

\newcommand{\beq}{\vspace{0mm}\begin{equation}}
\newcommand{\eeq}{\vspace{0mm}\end{equation}}
\newcommand{\beqs}{\vspace{0mm}\begin{eqnarray}}
\newcommand{\eeqs}{\vspace{0mm}\end{eqnarray}}
\newcommand{\barr}{\begin{array}}
\newcommand{\earr}{\end{array}}

\title{ImageBERT: Cross-modal Pre-training with Large-scale Weak-supervised Image-Text Data}

\author{
   Di Qi, Lin Su, Jia Song, Edward Cui, Taroon Bharti, Arun Sacheti \\
   Bing Multimedia Team, Microsoft \\
   \texttt{\{diqi,lins,jiaso,edwac,tbharti,aruns\}@microsoft.com} \\
}

\begin{document}

\maketitle

\begin{abstract}
In this paper, we introduce a new vision-language pre-trained model -- ImageBERT -- for image-text joint embedding. Our model is a Transformer\cite{2017arXiv170603762V}-based model, which takes different modalities as input and models the relationship between them. The model is pre-trained on four tasks simultaneously: Masked Language Modeling (MLM), Masked Object Classification (MOC), Masked Region Feature Regression (MRFR), and Image Text Matching (ITM). To further enhance the pre-training quality, we have collected a Large-scale weAk-supervised Image-Text (LAIT) dataset from Web. We first pre-train the model on this dataset, then conduct a second stage pre-training on Conceptual Captions\cite{Sharma2018ConceptualCA} and SBU Captions\cite{Ordonez2011Im2TextDI}. Our experiments show that multi-stage pre-training strategy outperforms single-stage pre-training. We also fine-tune and evaluate our pre-trained ImageBERT model on image retrieval and text retrieval\cite{2014arXiv1412.2306K} tasks, and achieve new state-of-the-art results on both MSCOCO\cite{2015arXiv150400325C} and Flickr30k\cite{young-etal-2014-image} datasets.
\end{abstract}

% % keywords can be removed
% \keywords{First keyword \and Second keyword \and More}

\section{Introduction}

Recently, vision-language tasks have attracted a lot of attention in both natural language processing (NLP) and computer vision (CV) communities. For example, Text-Image Retrieval\cite{2014arXiv1412.2306K} aims to retrieve the most relevant image given a text, or vice versa. Visual Question Answering (VQA)\cite{2015arXiv150500468A} aims to predict the correct answer given an image and an associated question. Visual Commonsense Reasoning (VCR)\cite{2018arXiv181110830Z} requires the model can not only answer the commonsense question but also select a rationale to support the answer. Image Captioning\cite{2017arXiv170707998A} aims to generate a natural language description for each input image. Based on pre-trained models trained by language and vision tasks separately (such as BERT\cite{2018arXiv181004805D} for language tasks and ResNet\cite{He2015DeepRL} for vision tasks), most previous methods used a late fusion manner to fuse the multi-modal inputs for downstream tasks. However, such late fusion layers usually require task-specific labeled data in training, but for many multi-modal tasks, acquiring enough task annotations is still very challenging and expensive.

Inspired by the success of pre-trained models in NLP, such as BERT\cite{2018arXiv181004805D}, XLNet\cite{2019arXiv190608237Y} and RoBERTa\cite{2019arXiv190711692L}, cross-modal pre-training has become a heated research area. Such models can learn joint representations for language and vision contents in the early stage based on large-scale corpus and then be applied to downstream tasks by task-specific fine-tuning. In this paper, We will first review  the latest work on cross-modal pre-training and compare their similarities and differences. Then, ImageBERT is proposed as a strong baseline for cross-modal pre-training, which has achieved new state-of-the-art results on text-to-image and image-to-text retrieval tasks on MSCOCO\cite{2015arXiv150400325C} and Flicker30k\cite{young-etal-2014-image}. We also build a new corpus, which includes 10M text-image pairs mined from web. We hope this corpus can further advance the development of cross-modal pre-training research.

%However, many recent work has demonstrated the effectiveness of Transformer-based model on these cross-modal tasks. We will review some of the recent works in the next section in details. Motivated by these great work, in this paper we propose our ImageBERT model which is also based on the Transformer architecture, and takes sub-word tokens as linguistic input together with densely sampled region-of-interest (RoI) image features as visual input. The model is pre-trained for multi-tasks, in order to model both the language and vision elements and their relationship. In addition, we have collected large-scale text-image training pairs from web using weakly supervised method, which is vital for improving our model performance on the downstream tasks. Our ImageBERT model has achieved new state-of-the-art results on image retrieval and text retrieval tasks on MSCOCO\cite{2015arXiv150400325C} and Flicker30K\cite{young-etal-2014-image}. Details will be introduced below.

\section{Related Work}

After Transformer\cite{2017arXiv170603762V} was proposed and widely used by cross-modal researches, the results on various tasks have been pushed to a new Everest in recent one year. Though almost all latest work are based on Transformer, they differ in various ways. We will review these work from different dimensions in below.
\begin{itemize}

\item \textbf{Model architecture.} BERT\cite{2018arXiv181004805D} model is pre-trained for NLP tasks whose input is one or two sentences. To apply BERT structure to cross-modal tasks, there can be many ways to deal with different modalities. ViLBERT\cite{Lu2019ViLBERTPT} and LXMERT\cite{Tan2019LXMERTLC} applied a single-modal Transformer to image and sentence respectively, then combined the two modalities together with a cross-modal Transformer. Other work, such as VisualBERT\cite{Li2019VisualBERTAS}, B2T2\cite{Alberti2019FusionOD}, Unicoder-VL\cite{2019arXiv190806066L}, VL-BERT\cite{2019arXiv190808530S}, Unified VLP\cite{Zhou2019UnifiedVP}, UNITER\cite{Chen2019UNITERLU}, etc., all concatenated image and sentence as a single input to the Transformer. It is hard to argue which model structure is better, since its performance really depends on the specific scenario.

\item \textbf{Image visual tokens.} Almost all recent paper applied an object detection model to the images and treated the detected regions of interest (RoIs) as image descriptors, just as linguistic tokens. Different from other work which used a pre-trained detection model, VL-BERT trained the detection network together with its image-text joint embedding network, and it also added global image features into model training. We can see region-based image features are good image descriptors, and they form a sequence of visual tokens that can be directly fed into Transformer.

\item \textbf{Pre-train data.} Unlike language model pre-training that can leverage tremendous natural language data, vision-language tasks require high quality image descriptions that are hard to obtain for free. Conceptual Captions\cite{Sharma2018ConceptualCA} is the most widely used data for image-text pre-training, given that it has 3M image descriptions and is relatively larger than other datasets. UNITER\cite{Chen2019UNITERLU} combines four datasets (Conceptual Captions\cite{Sharma2018ConceptualCA}, SBU Captions\cite{Ordonez2011Im2TextDI}, Visual Genome\cite{Krishna2016VisualGC} and MSCOCO\cite{2015arXiv150400325C}) together to form a 9.6M training corpus and achieved state-of-the-art results on many image-text cross-modal tasks. LXMERT\cite{Tan2019LXMERTLC} added some VQA training data into pre-training and obtained state-of-the-art results on VQA task. We can see that data quality and volume play important roles in model training, and should be paid more attention to when designing new models.

\end{itemize}

\section{Large-Scale Weak-supervised Image-Text Data Collection}
\label{sect:data-info}

\begin{figure*}[h]
    \centering
    \includegraphics[width=0.8\textwidth, height=18cm]{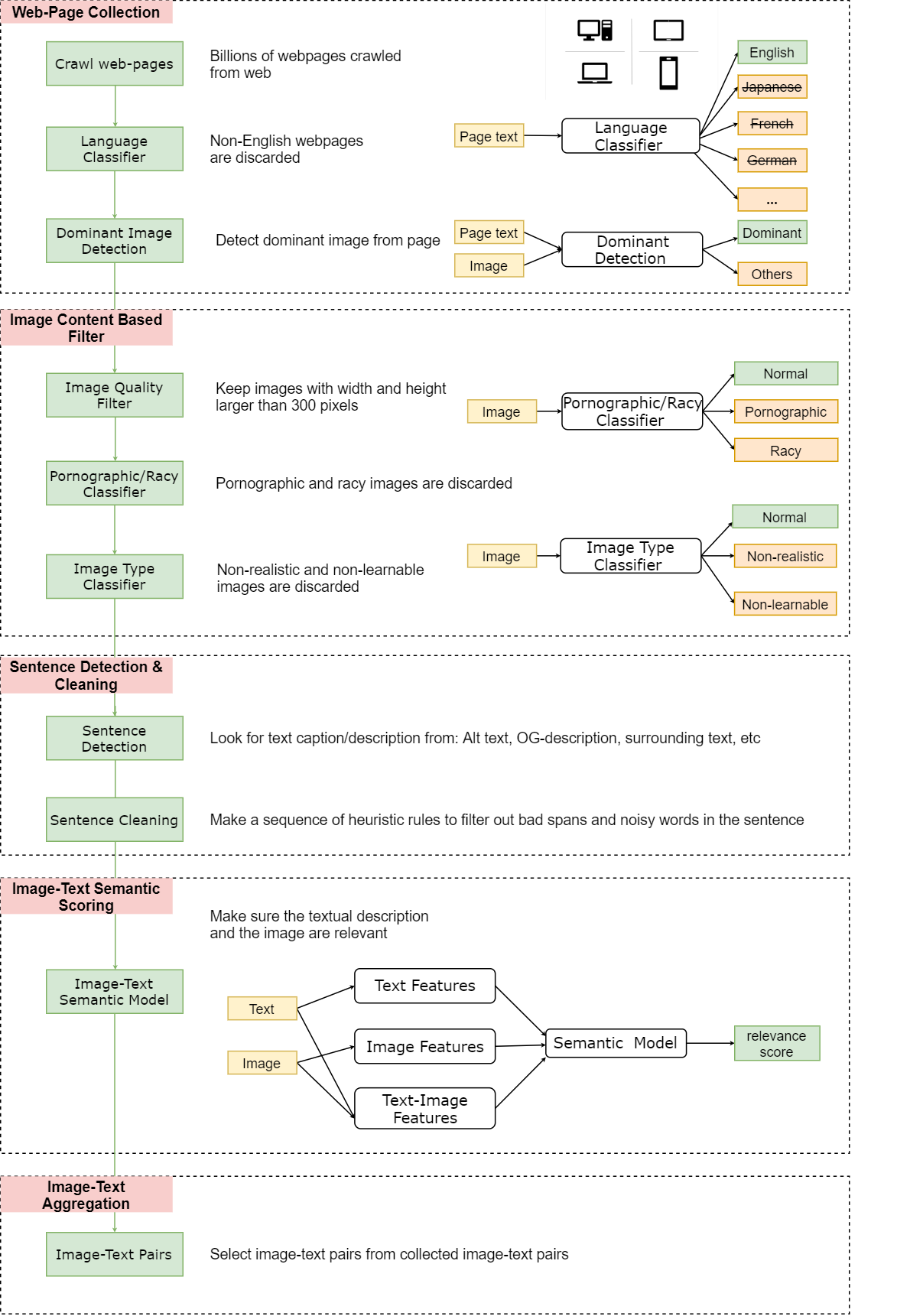}
    \caption{Weakly-supervised data collection pipeline}
    \label{fig:data_collection_pipeline}
\end{figure*}

Unlike language model based pre-training, which can use unlimited natural language texts, such as BooksCorpus\cite{Zhu_2015_ICCV} or Wikipedia, cross-modal pre-training requires large volume and high quality vision-language pairs. For example, most recent cross-modal pre-trained models \cite{Li2019VisualBERTAS,Alberti2019FusionOD,2019arXiv190806066L,2019arXiv190808530S,Zhou2019UnifiedVP,Chen2019UNITERLU} use following 2 datasets in pre-training: The Conceptual Captions (CC) dataset\cite{Sharma2018ConceptualCA}, which contains 3M images with descriptions harvested from the Alt-text HTML attribute of the web pages, and SBU Captions\cite{Ordonez2011Im2TextDI}, which consists of 1M images with user-associated captions. However, the size of these datasets is still not enough for pre-training a model which has several hundreds of million parameters, or even larger models in the future. In addition, image descriptions manually written by human can be of high-quality yet expensive. But there are innumerable web-pages on the Internet with associated images.

\begin{figure}[t]
    \centering
    \includegraphics[width=0.9\textwidth]{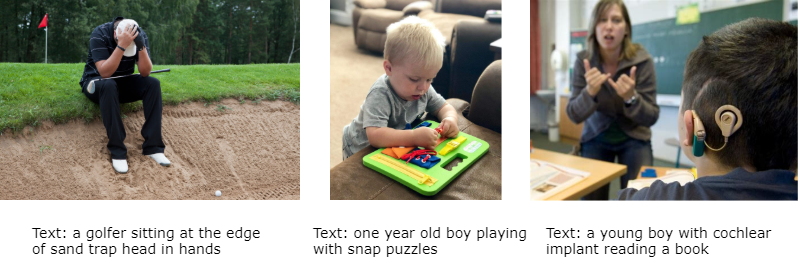}
    \caption{Data samples from LAIT dataset.}
    \label{fig:data_sample}
\end{figure}

Motivated by this, this paper designs a weak-supervised approach (as illustrated in Figure \ref{fig:data_collection_pipeline}) to collect large-scale image-text data from Web, whose volume and quality are essential to vision-language pre-train tasks. The resulting dataset LAIT (Large-scale weAk-supervised Image-Text) contains 10M images along with their descriptions with an average length of 13 words.
We will show in experiments that LAIT is beneficial for vision-language pre-training. Figure \ref{fig:data_sample} gives some examples. We will explain the data collection methodology in below.

\textbf{Web-page Collection.} We crawl billions of web-pages from Web and discard all non-English ones, given that all downstream tasks are in English. And then we parse each web-page to collect image URLs and detect dominant images by HTML tags and DOM tree features. Non-dominant images are discarded because they are more likely to be unrelated to the web-page content.

\begin{figure}[t]
    \centering
    \includegraphics[width=0.9\textwidth]{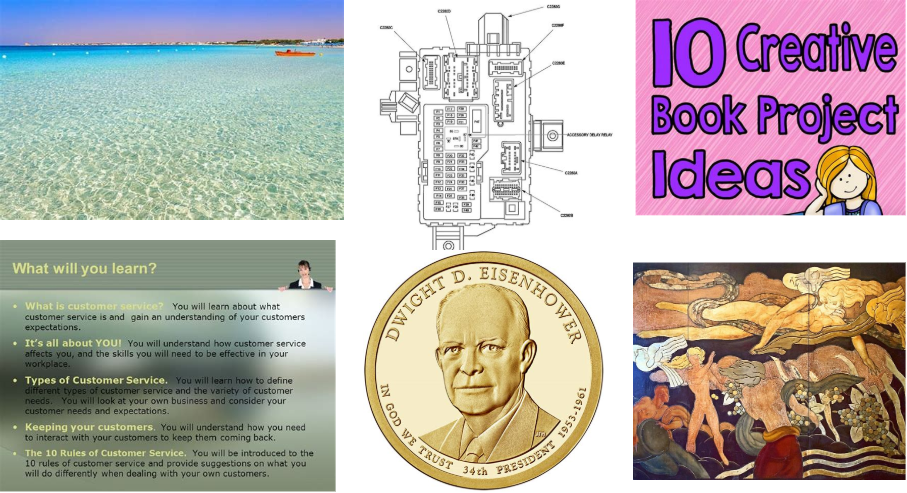}
    \caption{Examples of non-eligible images that have been discarded during data cleaning.}
    \label{fig:filter_data}
\end{figure}

\textbf{Image Content Based Filtering.}
We further filter the data based on image content. We only keep images whose width and height are both larger than 300 pixels. Also, images that contain pornographic or racy content are also discarded. Besides, given that the images in downstream tasks are all natural, realistic pictures taken from the real world, we apply a binary classifier to discard unnatural, non-realistic and non-learn-able images. Figure \ref{fig:filter_data} shows some examples of the non-eligible images that have been discarded during this process.

\textbf{Sentence Detection \& Cleaning.} We use below data sources as textual descriptions of the images: user-defined metadata in the HTML such as Alt or Title attribute, surrounding text of the image, etc.; we make a sequence of heuristic rules to filter out bad spans and noisy words (spam/pornography) in the sentences, and just keep sentences within a normal length. Finally we discard sentences that have high ratio of out-of-vocabulary terms.

\textbf{Image-Text Semantic Scoring.} After filtering bad images and cleaning up noisy text, we want to make sure the text and image are semantically relevant. With small supervised image-text data, we have trained a weak image-text semantic model to predict whether the $<text,image>$ pair is semantically relevant or not, and then apply it to the billion scale image-text pairs to filter out irrelevant pairs. The semantic model is trained upon hundreds of features including text-only features, image-content features, and text-image cross-modal features. 

\textbf{Image-Text Aggregation.} There are cases where one image is downloaded from multiple web-pages and thus has different text descriptions. In this case we only select the best scored $<text,image>$ pair. If too many images have the same description, we will directly drop all these pairs from the corpus.

\section{ImageBERT Model}
\label{sec:ImageBERT}

\begin{figure*}[t]
    \centering
    \includegraphics[width=1.0\textwidth]{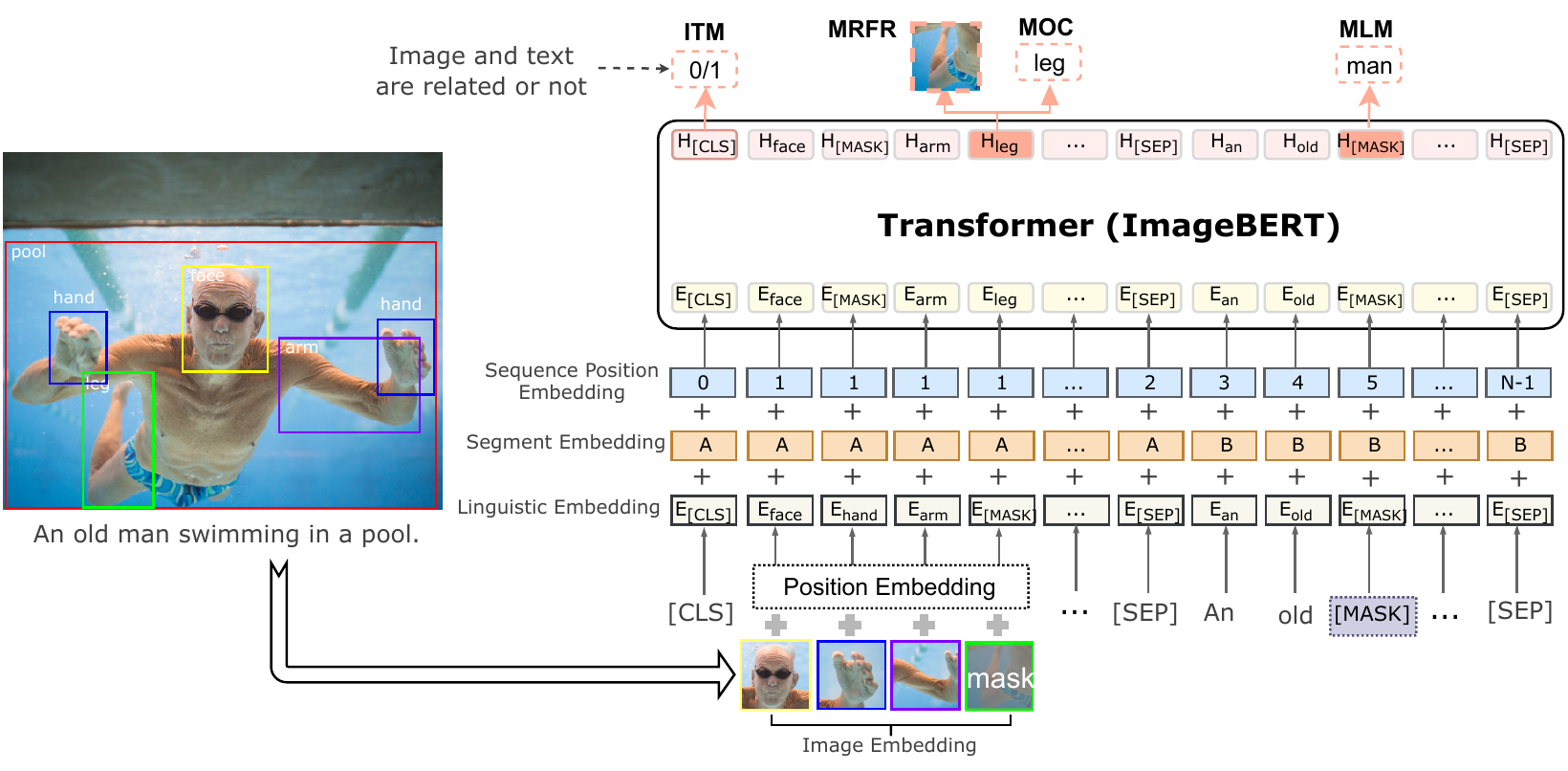}
    \caption{Architecture of ImageBERT model.}
    \label{fig:structure}
\end{figure*}

Figure \ref{fig:structure} illustrates the overall architecture of our ImageBERT model. Similar to BERT\cite{2018arXiv181004805D}, we use Transformer as basic structure, but take both image visual tokens and textual tokens as input. The image and text input are encoded into different embeddings through an embedding layer, where the image visual tokens are RoI features extracted from a Faster-RCNN\cite{singh2019TowardsVM,singh2018pythia} model. Then these embeddings are fed into a multi-layer bidirectional self-attention Transformer to learn a cross-modality Transformer to model the relationship between the visual regions and the linguistic tokens. 

\subsection{Embedding Modeling}
\label{sect:embedding}

We first introduce how we prepare the input for the Transformer through embedding layers.

\textbf{Linguistic embedding.} We adopt the similar word pre-prossessing method as BERT. The input sentence is tokenized into $n$ sub-word tokens \(\{w_0,\cdots,w_{n-1}\}\) using WordPiece\cite{2016arXiv160908144W} approach. Special tokens such as [CLS] and [SEP] are also added to the tokenized text sequence. The final embedding for each sub-word token is generated by combining its original word embedding, segment embedding and sequence position embedding (which will explained in details later). All these embeddings are initialized from public pre-trained BERT model.

\textbf{Image embedding.} Similar to linguistic embedding, image embedding is also generated from visual input by a similar process. A Faster-RCNN model is used to extract features from $o$ RoIs, denoted by \(\{r_0,\cdots,r_{o-1}\}\), from the image to represent its visual content. The detected objects can not only provide visual contexts of the whole image for the linguistic part, but also be related to specific terms through detailed region information. We also add position embeddings to image embeddings by encoding the object location with respect to the global image into a 5-D vector \( \textbf{c}^{(i)}=(\frac{x_{tl}}{W},\frac{y_{tl}}{H},\frac{x_{br}}{W},\frac{y_{br}}{H},\frac{(x_{br}-x_{tl})(y_{br}-y_{tl})}{WH}) \), where \((x_{tl},y_{tl})\) and \((x_{br},y_{br})\) denote top-left and bottom-right coordinates of the bounding box of the object respectively, and \(\frac{(x_{br}-x_{tl})(y_{br}-y_{tl})}{WH}\) denotes the proportional area with respect to the whole image. Both object features and location embeddings are projected into the same dimension with linguistic embeddings. The final representation $\textbf{e}^{(i)}$ for each image RoI $r^{(i)}$ is obtained via summing up its object embedding \(\textbf{v}^{(i)} = ImageEmbed(r^{(i)})\), segment embedding \(\textbf{s}^{(i)} = SegmentEmbed(i)\), image postion embedding \(\textbf{p}_{img}^{(i)} = PositionEmbed(\textbf{c}^{(i)})\) and sequence position embedding \(\textbf{p}_{seq}^{(i)} = PositionEmbed(i)\):
\[ \textbf{e}^{(i)} = LN(\textbf{v}^{(i)}+\textbf{s}^{(i)}+\textbf{p}_{img}^{(i)}+\textbf{p}_{seq}^{(i)})  \]
Each embedding is projected to a vector with the same embedding size as the hidden size in Transformer sub-layers, followed by Layer Normalization (LN).
We also use the classification label of each region from the detection model in a label prediction task (which will be explained in section~\ref{sect:tasks}). During our ablation study, we also experiment on adding global image features in addition to the region features.

\textbf{Sequence position and segment embedding.} Sequence position embedding for each token is used for indicating the order of the input tokens. We use a fixed dummy position for all the visual tokens, because there is no order of the detected RoIs, and the coordinates of the objects have already been added into the image embeddings. For the linguistic part, an ascending sequence is used to indicate the order of words in the textual description. Moreover, segment embedding is added to each input token to distinguish different modalities.

\subsection{Multi-stage Pre-training}
\label{sect:multi-stage}

\begin{figure*}[t]
    \centering
    \includegraphics[width=1.0\textwidth]{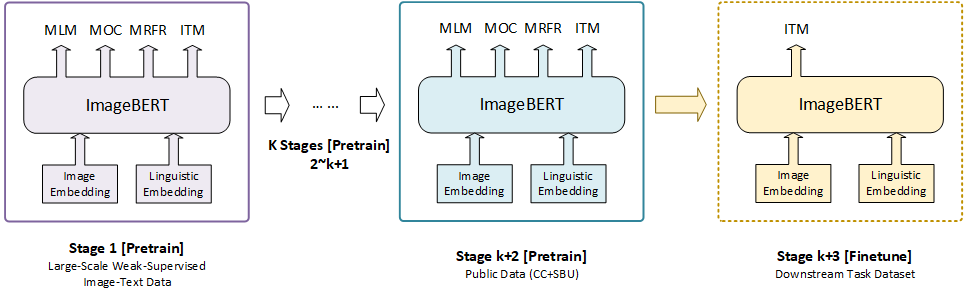}
    \caption{Multi-stage pre-training framework.}
    \label{fig:multi-stage}
\end{figure*}

Since different datasets are collected from different sources, they may have different quality levels and noise distributions. To better utilize different kinds of pre-train data, we propose a multi-stage pre-training framework, as shown in Figure \ref{fig:multi-stage}. According to downstream tasks, the pre-trained model should be trained firstly using large-scale out-of-domain data followed by small-scale in-domain data, so that the model can be better converged towards the final tasks. In multi-stage pre-training, several pre-train stages (k+2 stages, for example, in Figure \ref{fig:multi-stage}) could be applied to the same network structure to utilize different kinds of dataset sequentially. Unlike training strategy mentioned in \cite{Howard2018UniversalLM} which only has a single-stage in language model pre-training, language model fine-tuning and classifier fine-tuning separately, our multi-stage framework mainly applies to the pre-training stage so as to better utilize heterogeneous out-of-domain datasets. Another work \cite{Gong2012MultistageMF} which also mentioned the multi-stage concept used it to solve optimization problem in feature learning, which is very different from our multi-stage strategy here.

More specifically, in our ImageBERT model, we use a two-stage pre-training strategy. The first pre-training stage uses our LAIT dataset as mentioned in section~\ref{sect:data-info}, and the second stage uses other public datasets, such as Conceptual Captions and SBU Captions. Both pre-training stages use the same training strategy which includes all of our four pre-training tasks. We also conduct experiments on single-stage pre-training, which trains on all datasets simultaneously, but find it not work as well as multi-stage pre-training. The final fine-tuning stage uses the same model and parameters from first two stages, but discards all the tasks with masked terms or maked objects. 

During ablation study, we also experiment on different fine-tuning objectives for image-text retrieval tasks. We will introduce our pre-train tasks as well as fine-tune tasks in below.

\subsection{Pre-training tasks}
\label{sect:tasks}

During model pre-training, we design four tasks to model the linguistic information and visual content, as well as their interactions.

\textbf{Task 1: Masked Language Modeling (MLM). }
This task is the same with the MLM task in BERT\cite{2018arXiv181004805D} training. We denote the $n$ input sub-word tokens as \(\textbf{w}=\{w_0,\cdots,w_{n-1}\}\). The input token which will be predicted afterwards is masked randomly with a probability of 15\%. We denote the masked indices as \(\textbf{m}_T \in \mathbb{N}^M\) and the masked word as $w_{m_T}$, and the non-masked word as \(w_{\backslash m_T}\). The masked token is replaced with a special token [MASK], a random token or remains unchanged with a probability of 80\%, 10\%, 10\%, respectively. The prediction is made based on the surrounding text as well as the cross-attention between the textual tokens and the visual features, which are denoted as \(\textbf{v}=\{v_0,\cdots,v_{o-1}\}\), by minimizing the negative log-likelihood:
\[\mathcal{L}_{MLM}(\theta)=-E_{(\textbf{v},\textbf{w})\sim D}\log{P_{\theta}(w_{m_T}|w_{\backslash m_T},\textbf v)}\]
where $\theta$ denotes the model's parameters which are trainable and will be updated during training. The loss is calculated over all \((\textbf{v},\textbf{w})\) pairs in the training corpus $D$. The similar expression is omitted in sections below.

\textbf{Task 2: Masked Object Classification (MOC). }
This task is an expansion of the MLM task. Similar to language modeling, we also conduct masked modeling on the visual object tokens. We randomly mask each object token with a probability of 15\%, then zero out the masked token or keep the original token with a probability of 90\% and 10\%, respectively. We denote the masked indices of the object tokens as \(\textbf{m}_I \in \mathbb{N}^M\), the masked token as $v_{m_I}$, and non-masked token as \(v_{\backslash m_I}\). The $i^{th}$ of the $M$ masked input tokens is denoted as $v_{m_I}^{(i)}$. To predict the $M$ masked object tokens, we treat the labeled classification categories from Faster R-CNN model as ground truth labels \(l_{\theta}(v_{m_I}^{(i)})\). Suppose the output vector corresponding to the masked token from the Transformer is \(f_{\theta}(v_{m_I}^{(i)}) \in \mathbb{R}\), we add a fully-connected layer to predict the correct label from $K$ object classes, using the cross-entropy (CE) loss as final objective with the context of the linguistic features \(\textbf{w}\):
\[\mathcal{L}_{MOC}(\theta)=-E_{(\textbf{v},\textbf{w})\sim D}\sum_{i=0}^{M-1}CE(l_{\theta}(v_{m_I}^{(i)}),f_{\theta}(v_{m_I}^{(i)}))\]

\textbf{Task 3: Masked Region Feature Regression (MRFR).}
Similar to MOC, MRFR also models the visual content, but it does a more precise job on object feature prediction. This task aims to regress the embedding feature of each masked object, denoted as \(h_{\theta}(v_{m_I}^{(i)})\). We add a fully-connected layer on top of the output feature vector to project it to the same dimension with the pooled input RoI object feature, denoted as  \(r_{\theta}(v_{m_I}^{(i)})\). Then an L2 loss is applied to regress the ground truth feature:
\[\mathcal{L}_{MRFR}(\theta)=-E_{(\textbf{v},\textbf{w})\sim D}\sum_{i=0}^{M-1}\lVert h_{\theta}(v_{m_I}^{(i)})-r_{\theta}(v_{m_I}^{(i)})\rVert_2^2\]
Note that we use conditional mask for all of the above three tasks, which means we only calculate the all mask losses when the input image and text is related.

\textbf{Task 4: Image-Text Matching (ITM).}
In addition to the language modeling task and the visual content modeling tasks, we also add ITM task to learn the image-text alignment. For each training sample, We randomly sample negative sentences for each image and randomly sample negative images for each sentence, to generate negative training data. Therefore, we denote the ground truth label as \(y\in{\{0,1\}}\) for each image-text pair \((\textbf{v},\textbf{w})\) , indicating if the input sample pair is related or not.  Similar to BERT, we add [CLS] as the first token of the input sequence to ImageBERT model, and apply a fully-connected layer on top it to obtain the image-text similarity score \(s_\theta(\textbf{v},\textbf{w})\).
A binary classification loss is used for optimization:
\[\mathcal{L}_{ITM}(\theta)=-E_{(\textbf{v},\textbf{w})\sim D}[y\log{s_\theta(\textbf{v},\textbf{w})}+(1-y)\log{(1-s_\theta(\textbf{v},\textbf{w}))}]\]

\subsection{Fine-tuning tasks}
\label{sect:ft-task}
After pre-training, we get a well-trained pre-trained model on vision-language joint representation. We further fine-tune and evaluate this model on Image-Text Retrieval task. This task contains two sub-tasks: image retrieval and text retrieval. Image retrieval targets to retrieve the correct images given an input caption sentence describing the content of image. Text retrieval does a similar task in the opposite direction. We fine-tune on MSCOCO and Flickr30k dataset after a two-stage pre-training. During fine-tuning, the format of the input sequence is the same with that in pre-training but without any masks on the object or word. We propose two fine-tuning objectives  corresponding to different negative sampling methods: image-to-text (sample negative sentences for each image) and text-to-image (sample negative images for each text). Besides, we experiment on three different losses to get the best model quality:
\begin{itemize}
\item \textbf{Binary classification Loss.} This is to guarantee the prediction for negative samples is correct: the output scores of the negatives samples should not only be different from the positive sample but also be predicted with right labels. For a image-text pair \((\textbf{v},\textbf{w})\) with ground truth label \(y\in {{0,1}}\), we take the embedding of the first token \(t_{(\textbf{v},\textbf{w})}\) output from the Transformer as \(c_{\theta}(t_{(\textbf{v},\textbf{w})}) \in \mathbb{R}\), then apply the binary classification loss for optimization: 
\[\mathcal{L}_{BCE}(\theta)=-E_{(\textbf{v},\textbf{w})}[y\log{c_\theta(t_{(\textbf{v},\textbf{w})})}+(1-y)\log{(1-c_\theta(t_{(\textbf{v},\textbf{w})}))}]\]

\item \textbf{Multi-class Classification Loss}. This is the most widely-used loss to enlarge the margins between positive and negative samples. For each positive pair \((\textbf{v},\textbf{w})^+\), we sample $P-1$ negative pairs \((\textbf{v},\textbf{w})^-\) from different captions or images. Then add a scoring function to predict the correct label \(l_{(\textbf{v},\textbf{w})}\) for all $P$ pairs' first tokens \(t_{(\textbf{v},\textbf{w})}^{(j)}\), and apply the cross-entropy loss as final objective: 
\[\mathcal{L}_{CE}(\theta)=-E_{(\textbf{v},\textbf{w})}^{(j)}\sum_{j=0}^{P-1}CE(s(t_{(\textbf{v},\textbf{w})}^{(j)}),l_{(\textbf{v},\textbf{w})}^{(j)})\]

\item \textbf{Triplet Loss}. We use triplet loss to maximize the margin between positive and the hardest negative sample. As mentioned in multi-class classification loss, for each positive pair \((\textbf{v},\textbf{w})^+\), we sample $P-1$ negative pairs \((\textbf{v},\textbf{w})^-\) from different captions or images with calculated similarity scores \(s(t_{(\textbf{v},\textbf{w})^{-}})\). The hardest negative sample is given by \(n_h^-=\mathop{\arg\max}_{(\textbf{v},\textbf{w})^{j}\neq(\textbf{v},\textbf{w})^+}s(t_{(\textbf{v},\textbf{w})^{j}})\). Then we apply triplet loss to the positive sample and the hardest negative sample:
\[\mathcal{L}_{Triplet}(\theta)=-E_{(\textbf{v},\textbf{w})}^{(j)}\sum_{n^-\in\mathcal{N}}{\max[0,s(t_{(\textbf{v},\textbf{w})^{+}}),s(n_h^-)]}\]
\end{itemize}

We conduct ablation study on combinations of the above-mentioned three fine-tuning losses, and we will present experimental results later.

\section{Experiments}
For the Image-Text retrieval task, we present zero-shot result for evaluating the quality of the pre-trained model as well as the results after further fine-tuning. We compare our model with state-of-the-art methods on image-retrieval and text-retrieval tasks in different settings on MSCOCO\cite{2015arXiv150400325C} and Flickr30k\cite{young-etal-2014-image} datasets.

\subsection{Evaluation for the Pre-trained Model}
\textbf{Pre-train data.} We first pre-train our model on the LAIT dataset mentioned in section~\ref{sect:data-info}, with parameter initialized from the BERT-base model as stage-1. (Note that we only sampled 2M from LAIT for pre-training due to resource limitations, and are working on models with the complete 10M dataset.) Then we continue pre-training on public datasets (Conceptual Captions\cite{Sharma2018ConceptualCA} and SBU Captions\cite{Ordonez2011Im2TextDI}), as stage-2.

\textbf{Experiment settings.} Our model is a 12-layer Transformer with 768 hidden units, 3072 intermediate units, and 12 attention heads. We set dropout probability to 0.1, and use GELU as activation function. The max length of our input sequence is fixed to 144, which includes 100 visual tokens extracted from a Faster RCNN model pre-trained on Visual Genome dataset with 1600 categories, and other linguistic tokens and special tokens.

During pre-training, we use a batch size of 48 and a learning rate of 1e-4 with Adam optimizer. We train the model for 17 epochs using 4 V100 GPUs. Also, we use conditional mask in MLM, MOC and MRFR tasks, and only calculate the masked loss when the input pair is a positive sample.

We use the same evaluation metrics $R@K$ ($K=1,5,10$) as other work, which measure the percentage of correctly matched pairs in the top $K$-ranked results. Since both Flickr30k and MSCOCO contain five captions per image, sentence retrieval task is easier and can get higher scores than image retrieval task.

\begin{table*}
\small
 \centering
  \begin{tabular}{lcccccccccccc}
    \toprule
        \multirow{3}{*}{Method} &
        \multicolumn{6}{c}{Flickr30k} &
        \multicolumn{6}{c}{MSCOCO} \\
        & \multicolumn{3}{c}{Image Retrieval} &
        \multicolumn{3}{c}{Sentence Retrieval} &
        \multicolumn{3}{c}{Image Retrieval} &
        \multicolumn{3}{c}{Sentence Retrieval} \\
        & {R@1} & {R@5} & {R@10} & {R@1} & {R@5} & {R@10} & {R@1} & {R@5} & {R@10} & {R@1} & {R@5} & {R@10}\\
        \midrule
        \multicolumn{13}{c}{1k Test set} \\
        \midrule
        ViLBERT\cite{Lu2019ViLBERTPT} & 31.9 & 61.1 & 72.8 & - & - & - & - & - & - & - & - & -  \\
        Unicoder-VL\cite{2019arXiv190806066L} & 48.4 & 76.0 & 85.2 & 64.3 & 85.8 & 92.3 & 43.4 & 76.0 & 87.0 & 54.4 & 82.8 & 90.6\\
        UNITER\cite{Chen2019UNITERLU} & \textbf{62.3} & \textbf{85.6} & \textbf{91.5} & \textbf{75.1} & \textbf{93.7} & \textbf{95.5} & - & - & - & - & - & - \\
        \midrule
        \textbf{ImageBERT} & 54.3 & 79.6 & 87.5 & 70.7 & 90.2 & 94.0 & \textbf{53.6} & \textbf{83.2} & \textbf{91.7} & \textbf{67.0} & \textbf{90.3} & \textbf{96.1} \\
        \midrule
        \multicolumn{13}{c}{5k Test set} \\
        \midrule
        \textbf{ImageBERT} & - & - & - & - & - & - & \textbf{32.3} & \textbf{59.0} & \textbf{70.2} & \textbf{44.0} & \textbf{71.2} & \textbf{80.4} \\
    \bottomrule
  \end{tabular}
\caption{\label{pt-zs-table} Zero-shot results of our pre-trained model on Flickr30k and MSCOCO test sets.}
\end{table*}

\textbf{Zero-shot result of pre-train model.} Under this setting, we evaluate our pre-trained model on Flickr30k and MSCOCO test sets without fine-tuning, to measure the quality of the pre-trained model. Zero-shot results are shown in Table~\ref{pt-zs-table}. We can see our pre-trained model has achieved new state-of-the-art on MSCOCO, but is worse than UNITER model. Note that here we only show the result of stage-2 pre-training, result of stage-1 pre-training can be found in Table~\ref{ablation-table-pt} which will be explained in our ablation study of the pre-train datasets.  

Besides our model, only a few other works reported the zero-shot results of their model, so we only include the other two works in table~\ref{pt-zs-table}. Unicoder-VL\cite{2019arXiv190806066L} only contains one pre-training stage using public out-of-domain dataset (Conceptual Caption and SBU Captions). UNITER\cite{Chen2019UNITERLU} added partial in-domain datasets (VG and MSCOCO) during pre-training stage, thus getting the highest zero-shot results. But we can also get comparable results after fine-tuning, and we will present them later (refer to table~\ref{ft-table-retrieval}). It is also worth noticing that comparing to latest results of other methods which only have one pre-training stage, our multi-stage pre-training strategy learns more useful knowledge during pre-training, and can consequently contribute to the fine-tuning stage on the downstream tasks.

\subsection{Evaluation for the Fine-tuned Model}
\textbf{Experiment settings.} After two-stage pre-training on LAIT and other public datasets (Conceptual Captions and SBU Captions), we apply our well-trained model to downstream task, Image-Text retrieval, and fine-tune it for the ITM task. Experiments are also conducted on both Flickr30k and MSCOCO. During fine-tuning, we use a batch size of 24 and a learning rate of 5e-5, and train for 130 epochs using 4 V100 GPUs.

\textbf{Fine-tuned results.} The final results after fine-tuning on retrieval task are shown in Table~\ref{ft-table-retrieval}. We can see that our model achieves new state-of-the-art on both Flickr30k and MSCOCO (both on 1k and 5k test sets) and outperforms all the other methods, which proves the effectness of our LAIT data and our multi-stage pre-training strategy for cross-modal joint learning.

\begin{table*}
\small
 \centering
  \begin{tabular}{lcccccccccccc}
    \toprule
        \multirow{3}{*}{Method} &
        \multicolumn{6}{c}{Flickr30k} &
        \multicolumn{6}{c}{MSCOCO} \\
        & \multicolumn{3}{c}{Image Retrieval} &
        \multicolumn{3}{c}{Sentence Retrieval} &
        \multicolumn{3}{c}{Image Retrieval} &
        \multicolumn{3}{c}{Sentence Retrieval} \\
        & {R@1} & {R@5} & {R@10} & {R@1} & {R@5} & {R@10} & {R@1} & {R@5} & {R@10} & {R@1} & {R@5} & {R@10}\\
        \midrule
        \multicolumn{13}{c}{1k Test set} \\
        \midrule
        SCAN\cite{Lee2018StackedCA} & 48.6 & 77.7 & 85.2 & 67.4 & 90.3 & 95.8 & 58.8 & 88.4 & 94.8 & 72.7 & 94.8 & 98.4 \\
        SCG\cite{Shi2019KnowledgeAS}  & 49.3 & 76.4 & 85.6 & 71.8 & 90.8 & 94.8 & 61.4 & 88.9 & 95.1 & 76.6 & 96.3 & 99.2 \\
        PFAN\cite{Wang2019PositionFA} & 50.4 & 78.7 & 86.1 & 70.0 & 91.8 & 95.0 & 61.6 & 89.6 & 95.2 & 76.5 & 96.3 & 99.0 \\
        ViLBERT\cite{Lu2019ViLBERTPT} & 58.2 & 84.9 & 91.5 & - & - & - & - & - & - & - & - & - \\
        UNITER\cite{Chen2019UNITERLU} & 71.5 & 91.2 & 95.2 & 84.7 & 97.1 & 99.0 & - & - & - & - & - & - \\
        Unicoder-VL\cite{2019arXiv190806066L} & 71.5 & 90.9 & 94.9 & 86.2 & 96.3 & 99.0 & 69.7 & 93.5 & 97.2 & 84.3 & 97.3 & 99.3 \\
        \midrule
        \textbf{ImageBERT} & \textbf{73.1} & \textbf{92.6} & \textbf{96.0} & \textbf{87.0} & \textbf{97.6} & \textbf{99.2} & \textbf{73.6} & \textbf{94.3} & \textbf{97.2} & \textbf{85.4} & \textbf{98.7} & \textbf{99.8} \\
        \midrule
        \multicolumn{13}{c}{5k Test set} \\
        \midrule
        SCAN\cite{Lee2018StackedCA} & - & - & - & - & - & - & 38.6 & 69.3 & 80.4 & 50.4 & 82.2 & 90.0 \\
        Unicoder-VL\cite{2019arXiv190806066L} & - & - & - & - & - & - & 46.7 & 76.0 & 85.3 & 62.3 & 87.1 & 92.8 \\
        UNITER\cite{Chen2019UNITERLU} & - & - & - & - & - & - & 48.4 & 76.7 & 85.9 & 63.3 & 87.0 & 93.1 \\
        \midrule
        \textbf{ImageBERT} & - & - & - & - & - & - & \textbf{50.5} & \textbf{78.7} & \textbf{87.1} & \textbf{66.4} & \textbf{89.8} & \textbf{94.4} \\
    \bottomrule
  \end{tabular}
\caption{\label{ft-table-retrieval} Results of fine-tuned model on Flickr30k and MSCOCO test sets.}
\end{table*}

\subsection{Ablation Studies}
We also perform ablation experiments on different combinations of pre-train datasets, presence of global visual features, different training tasks, etc., on Flickr3k, so as to study deeply on our model structure and training strategy.

\begin{table*}
 \centering
  \begin{tabular}{lcccccc}
    \toprule
    \multirow{2}{*}{Version} &
      \multicolumn{3}{c}{Image Retrieval} &
      \multicolumn{3}{c}{Sentence Retrieval} \\
      & {R@1} & {R@5} & {R@10} & {R@1} & {R@5} & {R@10} \\
      \midrule
    LAIT & 21.4 & 46.0 & 59.1 & 31.6 & 58.4 & 72.0 \\
    %LAIT v2.0 & 23.7 & 52.1 & 63.4 & 32.2 & 61.7 & 72.3 \\
    CC & 33.0 & 60.6 & 72.5 & 43.8 & 73.7 & 81.7 \\
    CC+SBU & 48.4 & 76 & 85.2 & 64.3 & 85.8 & 92.3 \\
    \midrule
    LAIT+CC+SBU & 46.4 & 73.7 & 83.4 & 63.4 & 86.8 & 92.2 \\
    LAIT$\,\to\,$(CC+SBU) & \textbf{54.3} & \textbf{79.6} & \textbf{87.5} & \textbf{70.7} & \textbf{90.2} & \textbf{94.0} \\
    \bottomrule
  \end{tabular}
\caption{\label{ablation-table-pt} Abalation study on combinations of different datasets on Flickr30k test set.}
\end{table*}

\textbf{Pre-train dataset.} We conduct pre-train experiments using combinations of different datasets. Results are shown in Table~\ref{ablation-table-pt}. CC stands for pre-training only on Conceptual Captions dataset, SBU stands for pre-training only on SBU Captions, LAIT+CC+SBU stands for pre-training using LAIT, Conceptual Caption dataset and SBU combined dataset, LAIT$\,\to\,$CC+SBU stands for pre-training using LAIT as stage-1, then continue pre-training using Conceptual Captions and SBU Captions as stage-2. We can see that using three different out-of-domain datasets in a multi-stage way achieves significantly better results than all the other settings.

% version V1.1 and V1.3 contains 2M weak-supervised data collected from web search engine but V1.3 using more training tricks like more negative sample methods. V5.0 contains a huge volumn of 14M data which got best zero-shot result. 

\textbf{Global image features.} It is worth noticing that the detected RoIs may not include all the information of the whole image. Thus, we also try to add global image features to the visual part. We use three different Convolutional Neural Networks (CNN) models (DenseNet\cite{Iandola2014DenseNetIE}, Resnet\cite{He2015DeepRL} and GoogleNet\cite{Szegedy2014GoingDW}) to extract global visual features from an input image, but find not all of the metrics have improvements. Results can be seen in part 1 of Table~\ref{ablation-table}.

\textbf{Pre-train loss.} We also add MRFR loss which is inspired by UNITER\cite{Chen2019UNITERLU} to our pre-training, and achieve huge improvement on zero-shot results as shown in part 2 of Table~\ref{ablation-table}. This implies that adding a harder task for better modeling the visual content can contribute to visual-textual joint learning.

\textbf{Number of objects (RoIs) from image.} To understand the importance of the visual part in our model, we perform experiments on different numbers of objects. We use the setting of 100 objects in all the experiments above to provide enough context of the input image for the pre-training tasks. In our model, the 100 RoIs are extracted using a Faster R-CNN model to obtain the top-100 ranked objects ordered by confidence scores from the detection network. Since some bounding boxes of the objects may have overlaps with each other or contain duplicate information, we also conduct experiments to see the impact of different numbers of objects. As we can see in part 3 of Table~\ref{ablation-table}, with less objects (same number of objects with ViLBERT\cite{Lu2019ViLBERTPT}) our model gets no better results on retrieval task. We can conclude that more objects indeed can help the model achieve better results, because more RoIs are helpful for understanding the image content.

\textbf{Fine-tune loss.} For the three losses we mentioned in section~\ref{sect:ft-task}, we try different combinations of them during fine-tuning. As shown in part 4 of Table~\ref{ablation-table}, using binary cross-entropy loss itself gives the best fine-tuned results on image-text retrieval task.

\begin{table*}
 \centering
  \begin{tabular}{clcccccc}
    \toprule
    \multirow{2}{*}{ } &
    \multirow{2}{*}{Method} &
      \multicolumn{3}{c}{Image Retrieval} &
      \multicolumn{3}{c}{Sentence Retrieval} \\
      & & {R@1} & {R@5} & {R@10} & {R@1} & {R@5} & {R@10} \\
      \midrule
    \multicolumn{8}{c}{Pre-train zero-shot result} \\
    \midrule
    \multirow{2}{*}{1} &
    CC+SBU & 48.4 & \textbf{76.0} & \textbf{85.2} & \textbf{64.3} & 85.8 & 92.3 \\
    & CC+SBU(w global feature) & \textbf{49.2} & 75.9 & 84.9 & 64.0 & \textbf{87.4} & \textbf{92.4} \\
    \midrule
    \multirow{2}{*}{2} &
    CC & 33.0 & 60.6 & 72.5 & 43.8 & 73.7 & 81.7 \\
    & CC(w MRFR loss) & \textbf{34.7} & \textbf{61.7} & \textbf{73.4} & \textbf{45.3} & \textbf{73.8} & \textbf{84.2} \\
    \midrule
    \multicolumn{8}{c}{Fine-tune result} \\
    \midrule
    \multirow{2}{*}{3} & 
    RoI(36) & 68.7 & 88.6 & 92.3 & 84.3 & 96.5 & 97.9 \\
    & RoI(100) & \textbf{73.0} & \textbf{92.2} & \textbf{95.8} & \textbf{87.3} & \textbf{98.0} & \textbf{99.5} \\
    \midrule
    \multirow{4}{*}{4} & 
    Binary+CE+Triplet & 71.9 & 91.1 & 94.1 & 86.6 & \textbf{97.6} & 98.7 \\
    & CE only & 70.6 & 91.4 & 95.7 & 85.6 & 97.4 & \textbf{99.2} \\
    & Triplet only & 70.6 & 91.4 & 95.9 & 86.8 & 97.3 & 98.8 \\
    & Binary only & \textbf{73.1} & \textbf{92.6} & \textbf{96.0} & \textbf{87.0} & \textbf{97.6} & \textbf{99.2} \\
    \bottomrule
  \end{tabular}
\caption{\label{ablation-table} Ablation study on global image features, pre-train loss, number of RoIs, and fine-tune loss on Flickr30k test set.}
\end{table*}

\section{Conclusion}
In this paper, we presented a new vision-language pre-trained model, ImageBERT, which is based on Transformer architecture and models vision-language joint embedding. We also collected a large-scale image-text training corpus LAIT from Web using weakly-supervised methods, which has the largest volume among current existing vision-language datasets, and has demonstrated its effectiveness in the first stage of our multi-stage pre-training pipeline. We can see that large-scale out-of-domain data, though lack of precise human labels, can add value to the quality of the pre-trained model and consequentially benefit the corresponding downstream tasks. Our ImageBERT model has achieved new state-of-the-art results on both image retrieval and sentence retrieval tasks on MSCOCO and Flickr30k. In the future, we will try to extend our pre-trained model to other cross-modal tasks such as VQA, VCR, and image captioning.

\section{Acknowledgement}
This work was done with invaluable help from colleagues from Microsoft Research Asia. We thank Nan Duan, Gen Li, and Haoyang Huang (\{nanduan,haohua\}@microsoft.com, ligen.li@pku.edu.cn) for their great support and contribution to the source code, training data and this paper. The authors are also grateful to their insightful ideas and helpful discussions on the model design and training.

\bibliographystyle{unsrt}  
\bibliography{ref}
\end{document}